\documentclass[letterpaper, 10 pt, conference]{ieeeconf}

\IEEEoverridecommandlockouts
\usepackage{array}
\usepackage[T1]{fontenc}
\usepackage{cite}
\usepackage{amsmath,amssymb,amsfonts}
\usepackage{algorithmic}
\usepackage{graphicx}
\usepackage{textcomp}
\usepackage{xcolor}
\usepackage{caption}
\usepackage{algorithm,algorithmic}
\usepackage{lipsum}
\usepackage{gensymb}
\usepackage{mleftright}
\usepackage{bm}
\usepackage{soul}
\usepackage{comment}
\usepackage{mathrsfs}
\usepackage{multirow}
\usepackage{siunitx}
\let\labelindent\relax
\usepackage{enumitem}
\usepackage{microtype}
\usepackage{cleveref}
\usepackage{booktabs}
\usepackage{fvextra} 
\usepackage{adjustbox}
\setstcolor{red}
\usepackage{makecell}
\renewcommand{\arraystretch}{1.0}

\usepackage{tikz}
\usetikzlibrary{
  arrows.meta,
  positioning,
  backgrounds,
  calc,
  shadows.blur
}
\definecolor{navy}{RGB}{20,40,90}
\definecolor{deepblue}{RGB}{38,83,145}
\definecolor{softblue}{RGB}{231,240,250}
\definecolor{softgreen}{RGB}{233,244,233}
\definecolor{softamber}{RGB}{255,247,230}
\definecolor{amber}{RGB}{170,100,10}
\definecolor{certfill}{RGB}{185,210,238}
\definecolor{failred}{RGB}{170,35,35}
\definecolor{lightred}{RGB}{255,235,235}
\definecolor{darkgreen}{RGB}{35,110,60}

\usepackage{url}

\usepackage{tcolorbox}
\tcbuselibrary{skins,breakable}
\usepackage{ragged2e}     
\usepackage{microtype}    
\usepackage[T1]{fontenc}
\usepackage{underscore}   

\usepackage{amsthm}
\usepackage[algo2e,ruled,vlined,linesnumbered,resetcount,norelsize]{algorithm2e}

\usepackage[normalem]{ulem}

\theoremstyle{plain}

\theoremstyle{definition}

\usepackage{mathtools}

\theoremstyle{remark}

\makeatletter
\newcommand{\linebreakand}{%
  \end{@IEEEauthorhalign}
  \hfill\mbox{}\par
  \mbox{}\hfill\begin{@IEEEauthorhalign}
}
\makeatother

\theoremstyle{definition}

\theoremstyle{plain}

\def\BibTeX{{\rm B\kern-.05em{\sc i\kern-.025em b}\kern-.08em
    T\kern-.1667em\lower.7ex\hbox{E}\kern-.125emX}}
\begin{document}

\renewcommand{\baselinestretch}{.98}

\title{Contact as a Decision Variable: Capability-Tradeoff Contact Selection for Legged Loco-Manipulation}
\author{
Al Jaber Mahmud, Shuai Li, and Xuan Wang
\thanks{A. Mahmud and X. Wang are with George Mason University. S. Li is with the University of Florida. 
}
} 

\maketitle

\begin{abstract}

In this paper, we study the joint selection of an environmental support contact and a whole-body configuration for a prescribed loco-manipulation task. A contact may provide greater physical support while restricting the motion required for the task. We formulate this problem through three capability measures: residual wrench, end-effector reach, and base mobility available after satisfying the task requirements, and we balance them against contact acquisition cost. Evaluating these capabilities for every candidate requires repeated whole-body optimizations. To reduce this computational cost, we propose Capability-Tradeoff Contact Selection (CTCS). CTCS screens candidates for contact and task feasibility, groups similar candidates within each surface, and predicts their capabilities from exact anchor evaluations using local sensitivity analysis. It checks these predictions through selective exact evaluations, ranks candidates by capability, and evaluates a shortlist exactly for final selection. We evaluate CTCS in simulations and hardware experiments using a Unitree Go2 quadruped with an AgileX NERO arm across $392$ task conditions with nine available support surfaces. Results show that CTCS outperforms ground-only and fixed-contact support, as it can select support surfaces that provide favorable capability trade-offs for the task. Compared with evaluating every candidate exactly, CTCS achieves approximately $3\times$ speedup while closely matching the resulting mean objective value.

\end{abstract}

\section{Introduction}
\label{sec:introduction}
\begin{figure}[t]
\centering
\includegraphics[width=\columnwidth]{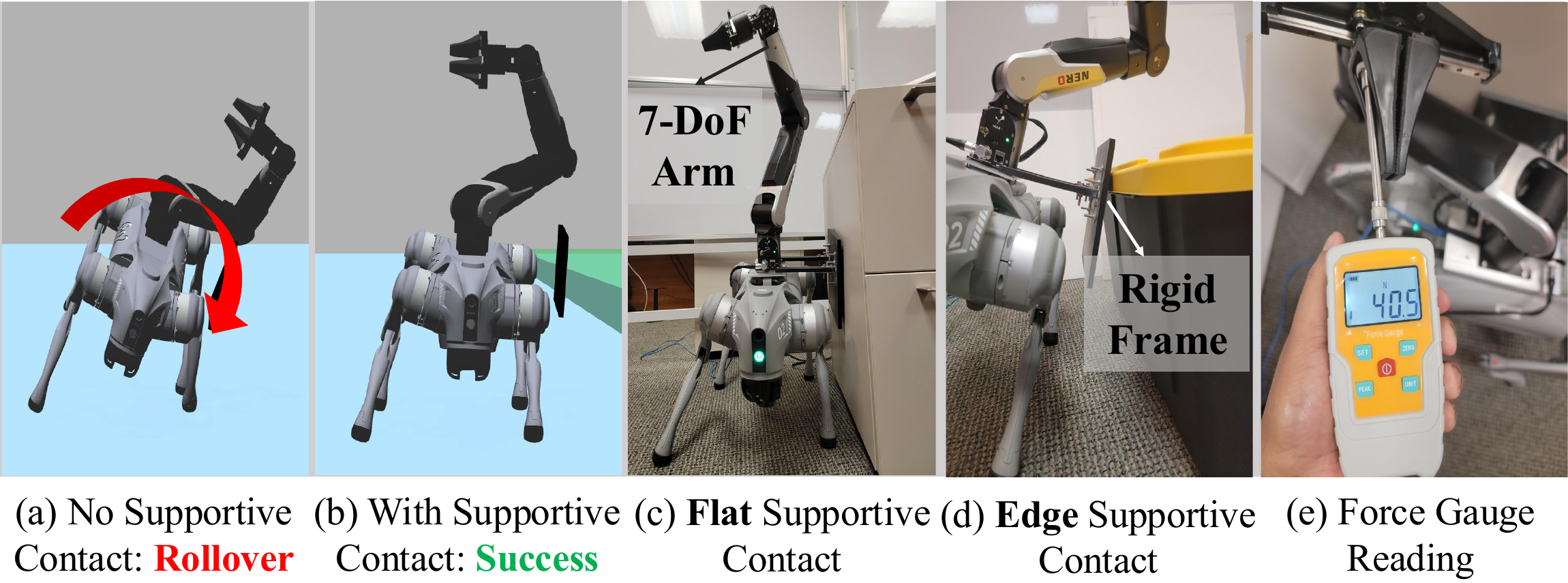}
\caption{\small Environmental support contact for a legged mobile manipulator. Under a large end-effector wrench, the robot (a) rolls over without environmental support and (b) sustains the wrench with support. Through contact selection, the robot establishes (c) a flat contact with one surface and (d) an edge contact with another surface. 
(e) End-effector force measurement.\vspace{-2.0em}}
\label{fig:concept_robot_pic}
\end{figure}

In legged locomotion and manipulation, the robot needs to balance itself and move while simultaneously performing manipulation tasks. This becomes challenging when the task requires long reaches or large external wrenches at the end effector.
In complex environments, surrounding structures can obstruct robot motion, but they can also provide useful physical support. Similar to humans that can lean against a table edge to reach deeper, legged mobile manipulators can also utilize environmental contact to perform tasks that would otherwise cause them to lose balance (e.g.,
Fig.~\ref{fig:concept_robot_pic}(a-b)). 
Furthermore, the same environment may provide multiple ways of establishing such support. As illustrated in
Fig.~\ref{fig:concept_robot_pic}(c-d), 
different contact realizations can lead to substantially different whole-body configurations and task capabilities.

In this paper, we consider the environmental support itself as a decision variable. For a prescribed loco-manipulation task, we jointly select a sustained auxiliary support contact from the available environmental surfaces and a whole-body configuration that realizes it. 
For example, consider a mobile manipulator walking while spraying a water
cannon. The robot needs to leverage environmental contact to sustain the reaction wrench generated by the water jet, reserve sufficient
end-effector motion to adjust the pose of the water cannon, and at the same time, the contact friction should not restrict the base's ability to continue walking.
This example illustrates the non-trivial trade-off in support selection: a realization that provides greater physical support may also impose stronger restrictions on the motion required for manipulation or locomotion. The desired contact therefore depends jointly on the load and motion requirements of the task. We
characterize this trade-off through the additional wrench, end-effector
motion, and base motion that remain available after the prescribed task
requirements are satisfied. For walking, these capabilities must additionally
remain adequate as the stance-foot support changes.

Environmental support has previously been exploited to improve manipulation
and whole-body performance~\cite{fang2019exploitation,
wolfslag2020optimisation}. Contact-planning methods aim to select footholds
or other environmental contacts as part of locomotion and trajectory
planning~\cite{tonneau2020sl1m,natarajan2023torque,sleiman2023versatile}.
In contrast, we consider contact selection: which sustained auxiliary support realization should be established for a given loco-manipulation task.
This selection becomes computationally challenging when many environmental surfaces and whole-body configurations are available. Evaluating the task-dependent capabilities of each candidate requires repeated whole-body optimization under task, contact, actuation,
and ground-support constraints. Exhaustive sampling or search may therefore
require many exact evaluations, while local sensitivity predictions may
become unreliable when active constraints change or when candidates belong
to different contact surfaces.

To address these challenges, we develop \emph{Capability-Tradeoff Contact Selection} (CTCS). The key idea is to separate exact verification of constraint satisfaction from the more expensive evaluation of candidate capabilities. Task and contact feasibility (less computationally expensive) are checked exactly, while the remaining capability evaluations (computationally expensive) are accelerated by grouping locally similar candidates, predicting their capabilities from exact anchor evaluations through local sensitivity analysis. The predictions are checked using
selective exact evaluations, and the most promising candidates are evaluated
exactly before the final contact realization is selected. To this end, our contributions are threefold:

\noindent$\bullet$ We formulate environmental support as a joint surface--configuration selection problem. Candidate supports must satisfy the task load and motion requirements, and are further optimized based on capability measures concerning residual wrench, end-effector reach, and base mobility.

\noindent$\bullet$ We develop CTCS, an efficient candidate-evaluation method that
    combines exact task-feasibility screening with local sensitivity-based
    capability prediction. Candidates are grouped within individual contact surfaces to enable reliable anchor-based prediction among locally similar realizations.

\noindent$\bullet$ We validate the proposed formulation and CTCS through
both simulation and hardware experiments, demonstrating that the selected
environmental contacts satisfy the prescribed task requirements while
providing favorable capability trade-offs.

\section{Literature Review}
\label{sec:related_work}

\subsection{Environmental Contacts for Whole-Body Support}
Legged mobile manipulators typically rely on stance-foot contacts for whole-body support~\cite{liu2025visual,molnar2025whole,rigo2024hierarchical,fu2023deep}.
Additional contacts with environmental structures have also been investigated for humanoid and legged robots~\cite{polverini2020multi, caron2015leveraging}, as hand and body contacts with environments can help humanoid robots maintain or recover balance under disturbances~\cite{wang2017real, murooka2024whole, henze2017multi}.
For manipulation, environmental bracing has been used to reduce joint effort, enable torque-limited manipulation tasks, and enlarge the workspace reachable under torque limits~\cite{natarajan2023torque}, while additional body-ground contacts have been used to improve quadruped robustness and free limbs for manipulation~\cite{wolfslag2020optimisation}. 
These works show how additional contacts can benefit robot task performance. However, even on the same environmental surface, different realizations can provide different wrench capabilities while imposing different restrictions on end-effector motion and base mobility. 
This motivates explicitly selecting the support realization according to the task.

\subsection{Contact Selection}
Contact selection has been studied jointly with locomotion and manipulation planning. Mixed-integer methods have been used to select terrain
regions, contact locations, and gaits~\cite{deits2014footstep,aceituno2017simultaneous}. SL1M~\cite{tonneau2020sl1m} uses a sparse $\ell_1$-norm relaxation to select among candidate terrain surfaces and determine footstep locations for successive steps. 
Multi-contact planning methods have been used to determine contact sequences and robot motions in complex environments~\cite{carpentier2018multicontact,tazaki2022fast}.
For legged loco-manipulation,~\cite{sleiman2023versatile} jointly determine contact schedules and whole-body trajectories to perform tasks involving interactions with the environment.
Natarajan et al.~\cite{natarajan2023torque} incorporate environmental bracing into manipulation
planning to enable tasks under joint-torque limits.

In contrast, we consider a given loco-manipulation task and jointly select a sustained auxiliary support surface and the whole-body configuration that realizes the contact.
This differs from works that assume a preselected support surface or support type and optimize how the robot uses that contact~\cite{fang2019exploitation,wolfslag2020optimisation}. It also differs from contact planning methods that optimize sequential footholds~\cite{deits2014footstep,aceituno2017simultaneous,tonneau2020sl1m} or environmental contacts~\cite{natarajan2023torque,tazaki2022fast} as part of locomotion or trajectory planning.

\subsection{Efficient Evaluation of Contact Candidates}

Evaluating task capabilities for many surface and configuration candidates can require repeated whole-body optimization under task, contact, and actuation constraints. Prior works have reduced contact planning computation through sparse optimization~\cite{tonneau2020sl1m}, search combined with trajectory optimization~\cite{tazaki2022fast,sleiman2023versatile}, and learned predictions of contact-transition feasibility and cost~\cite{lin2019efficient}. 
Differentiable optimization methods provide sensitivity information about how optimization solutions vary with problem parameters under regularity conditions~\cite{amos2017optnet,agrawal2019differentiable}.
However, our problem requires repeatedly evaluating task-dependent whole-body capabilities for many candidate support realizations spanning discrete surface choices with different geometric and contact constraints.
Even within the same surface, nearby configurations may have different active torque or friction constraints, so their optimized capability measures may not vary smoothly.
Consequently, exhaustive sampling or search may require many exact evaluations and become computationally costly, whereas local sensitivity predictions may become unreliable across changes in active constraints or contact surfaces.

\section{Preliminaries and Problem Formulation}
\label{sec:problem_formulation}
We consider a quadruped mobile manipulator that can use environmental surfaces as additional physical support during a loco-manipulation task. 
Our objective is to select an environmental contact and whole-body configuration that satisfy the task’s load and motion requirements while providing a favorable trade-off among the capabilities available under the task load. We first introduce the robot and contact models, then define the task requirements and capability measures, and finally formulate the joint selection problem.

\noindent\textbf{Preliminaries: Robot Whole-body Dynamics.} We consider the legged mobile manipulator illustrated in
Fig.~\ref{fig:concept_robot_pic}, 
which has an auxiliary surface mounted on the side of its body to facilitate contact with the environment. To model the robot,
we denote the robot configuration as $q=(p_b,R_b,q_\ell,q_m)$, where $p_b\in\mathbb R^3$ and $R_b\in\mathrm{SO}(3)$ are the base position and orientation, and $q_\ell$ and $q_m$ are the leg and manipulator joint angles. The corresponding generalized velocity is
$v= [v_b^\top ~ \omega_b^\top ~ \dot q_\ell^\top ~ \dot q_m^\top]^{\top}$, where $v_b$ and $\omega_b$ are the base linear and angular velocities. The whole-body dynamics are
\begin{equation}
M(q)\dot v+h(q,v)
\!=\!
S^\top u
+\!\sum_i \! J_{g,i}^{\top}(q)\lambda_{g,i}
+J_c^\top(q)\lambda_c
+J_e^\top (q) w_e
\label{eq:dynamics}
\end{equation}
where $M(q)$ is the inertia matrix, $h(q,v)$ contains the bias terms, $S$ is the actuation-selection matrix, and $u$ is the vector of actuated joint torques.
The forces $\lambda_{g,i}$ and $\lambda_c$ are the ground reaction force at stance foot $i$ and the environmental support force, respectively, and $w_e$ is the external wrench applied at the end effector. The matrices $J_{g,i}(q)$, $J_c(q)$, and $J_e(q)$ are the corresponding Jacobians. The summation runs over the stance feet. The exact definitions of the quantities in \eqref{eq:dynamics} are given in the Supplementary Material~\cite{supp}.

\subsection{Definition of Environmental Contacts}

The robot can use surrounding surfaces for physical support. Let $\mathcal E$ be the finite set of such planar environmental surfaces. Each surface is denoted by $\Pi= (p_\Pi,n_\Pi, \mathcal A_\Pi)\in\mathcal E$, where $p_\Pi\in\mathbb R^3$ is a reference point on the surface, $n_\Pi\in \mathbb R^3$ is the unit surface normal pointing toward the robot, and  $\mathcal A_\Pi\subset\mathbb R^3$ denotes the usable region of the surface (in the world frame).

\noindent\textbf{Contact realization conditions}:
We denote a candidate contact realization by $\chi=(q,\Pi)$, where $q$ is the whole-body robot configuration and $\Pi$ is the environmental contact surface.  
The following conditions are required to establish the contact on $\Pi$:\\
$\bullet$ \textbf{(C1)} $n_\Pi^\top(p_c(q)-p_\Pi)=0$ and $p_c(q)\in\mathcal A_\Pi$, where $p_c(q)$ is the reference contact point of the robot. \\
$\bullet$ \textbf{(C2)} Depending on whether the task involves locomotion of the mobile base, we define two \textit{contact modes}:
\begin{equation}
\sigma=\begin{cases}
\mathrm{sticking},& J_c v=0,\\
\mathrm{sliding},&n_\Pi^{\top}J_c v=0,~~ (I-n_\Pi n_\Pi^{\top})J_c v\neq 0.
\end{cases}
\label{eq:mode_map}
\end{equation}
where $J_c(q) v$ represents the linear velocity of the environmental contact point.
Sticking requires zero contact-point velocity; sliding allows nonzero tangential velocity. We use static-friction constraints for sticking and kinetic friction opposing the sliding velocity for sliding. The explicit friction models are provided in the Supplementary Material~\cite{supp}.

\subsection{Task Requirements and Capability Measures}

\noindent{\textbf{Task Requirements.}
We specify a task by a required end-effector wrench $w_e^d \in\mathbb R^6$, a required end-effector position $p_e^d \in\mathbb R^3$, and a required base velocity $v_b^d \in\mathbb R^3$ along a task-required direction. A candidate $\chi$ satisfies the task requirements if the whole-body constraints admit a feasible solution satisfying condition:\\
$\bullet$ \textbf{(T1)} 
$(w_e = w_e^d,~
p_e = p_e^d,~
v_b = v_b^d)$ holds, $\forall s\in\mathcal S$, \\
where $s\in\mathcal S$ represents the stance phases, including the two diagonal trot phases (two-leg support) that the robot alternates during walking and the four-foot stance for the sticking-mode manipulation task.

Candidates satisfying task requirement \textbf{(T1)} may still differ in their capacity to withstand additional disturbances and adjust their end-effector position or base velocity. To capture this, for each candidate $\chi$ and stance phase $s$, we evaluate
the feasible variation of one task quantity around its required value while maintaining the other task quantities at their required values. The three task quantities then yield residual wrench capability, residual end-effector reach capability, and residual base mobility, respectively:

\noindent\textbf{Residual Wrench Capability.}
We measure the additional disturbances that the robot can withstand while maintaining the required end-effector position $p_e^d$ and base velocity $v_b^d$.
Let $R_T \in SO(3)$ be the task-dependent rotation matrix that maps quantities from the robot frame to the task-aligned frame.
For wrench evaluation, we define 
$T_\ell = \begin{bsmallmatrix}
R_T & 0\\
0 & \ell_w^{-1}R_T
\end{bsmallmatrix}$, where $\ell_w>0$ scales force and moment components into comparable units.
The residual wrench set is\footnote{Here and below, \textit{feasible} means that there exist whole-body control variables satisfying the constraints listed above for the fixed candidate \(\chi\) and stance s, corresponding to the whole-body dynamics~\eqref{eq:dynamics}. The complete constraint sets are provided in the Supplementary Material~\cite{supp}. }
\begin{equation}
\mathcal W_{\mathrm{wrench}}^s(\chi)
=\bigl\{\delta w\in\mathbb R^6 \;\big|\;
w_e^d+T_\ell^{-1}\delta w
\text{ is feasible}\bigr\}.
\end{equation}
Here, $\delta w$ is the scaled wrench variation expressed in
the task-aligned frame. The origin ($\delta w=0$) corresponds to the required
wrench $w_e^d$, which must be feasible for every stance phase, i.e. ($0\in\mathcal W_{\mathrm{wrench}}^s$). 

Each axis direction $e_j \in \mathbb{R}^{6}$ is defined in the task-aligned frame, so that $T_\ell^{-1} e_j$ gives the corresponding direction in the original coordinate frame. For each $e_j$, let $r_{\mathrm{wrench},j}^{s,+}$ and $r_{\mathrm{wrench},j}^{s,-}$ denote the largest feasible residual magnitudes along $+e_j$ and $-e_j$, and define $r_{\mathrm{wrench},j}^s
=
\min\left\{
r_{\mathrm{wrench},j}^{s,+},
r_{\mathrm{wrench},j}^{s,-}
\right\}$. 
For a fixed robot state and fixed contact mode, $\mathcal W_{\mathrm{wrench}}^s$ contains the convex hull of the axis points $\{\pm r_{\mathrm{wrench},j}^s e_j \}_{j=1}^{6}$.
This hull provides a conservative inner approximation of the residual wrench region. We define the residual wrench capability as the radius of the largest origin-centered Euclidean ball~\cite{qiu2022new,orsolino2018application} contained in this hull, expressed in the scaled task-aligned coordinates:
\begin{equation}\textstyle
\rho_{\mathrm{wrench}}^s
=
\left(
\sum_{j=1}^{6}
\frac{1}
     {(r_{\mathrm{wrench},j}^s)^2}
\right)^{-1/2}.
\end{equation}

\noindent\textbf{Residual End-Effector Reach Capability.}
We measure the feasible variation of the end-effector position
around $p_e^d$ while sustaining the required wrench $w_e^d$
and base velocity $v_b^d$. The end-effector position $p_e$
and its required value $p_e^d$ are expressed in the same
reference frame. Define
\begin{equation}
\mathcal W_{\mathrm{reach}}^s(\chi)
\!=\!\bigl\{\delta p\in\mathbb R^3 \;\big|\;
\delta p=\tau R_T\dot p_e,\;
\dot p_e \text{ is feasible}\bigr\},
\end{equation}
where $\delta p$ is the position variation expressed in the task-aligned frame and $\dot p_e$ is the end-effector velocity expressed in the same reference frame, and $\tau>0$ is a fixed short horizon.
$\delta p=\tau R_T \dot p_e$ provides a local displacement measure.

Using a similar directional-reserve construction to the wrench capability, 
along the task-aligned position axes denoted by $r_{\mathrm{reach},j}^{s,+}$ and
 $r_{\mathrm{reach},j}^{s,-}$, 
the residual end-effector reach capability is
\begin{equation}\textstyle
\rho_{\mathrm{reach}}^s
=
\left(
\sum_{j=1}^{3}
\frac{1}
     {(r_{\mathrm{reach},j}^s)^2}
\right)^{-1/2}.
\end{equation}

\noindent\textbf{Residual Base Mobility.}
For walking tasks, we measure the feasible variation of base
velocity along the prescribed unit motion direction
$\frac{v_b^d}{|v_b^d|}$, tangent to the environmental surface.
We maintain $w_e=w_e^d$ and $p_e=p_e^d$ and vary the base velocity along the task direction to define
\begin{equation}\textstyle
\mathcal W_{\mathrm{mob}}^s(\chi)
=\bigl\{\delta\nu\in\mathbb R \;\big|\;
v_b^d+\delta\nu\,\frac{v_b^d}{|v_b^d|}
\text{ is feasible}\bigr\}.
\end{equation}
Since this directed velocity has an effective dimension of $1$, 
let $r_{\mathrm{mob},1}^{s,+}$ and
$r_{\mathrm{mob},1}^{s,-}$ denote the largest feasible
positive and negative velocity variations.
The residual mobility measure is
\begin{equation}\textstyle
\rho_{\mathrm{mob}}^s
=
\min\left\{
r_{\mathrm{mob},1}^{s,+},
r_{\mathrm{mob},1}^{s,-}
\right\}.
\end{equation}
For all capability measures, if any directional reserve is zero, we set the corresponding $\rho^s=0$.

We evaluate each capability using its weakest phase among $s\in \mathcal S$, and we define 
\begin{equation*}\textstyle
\underline\xi(\chi)=\big[\min_{s\in\mathcal S} \rho_{\mathrm{wrench}}^s ~ \min_{s\in\mathcal S} \rho_{\mathrm{reach}}^s ~ \min_{s\in\mathcal S} \rho_{\mathrm{mob}}^s\big]^\top.
\end{equation*}
For the sticking-mode manipulation task, the mobility component is omitted from $\underline\xi(\chi)$.

\subsection{Problem of Interest}
Our objective is to find a whole-body configuration and corresponding environmental contact that can satisfy task requirements and provide a favorable trade-off among the three capabilities.
Using the capability vector defined in Section~III-B, we define the capability utility as
\begin{equation*}  \textstyle
U(\underline\xi(\chi))=\sum_{k\in\mathcal K_{\mathrm{task}}} \alpha_k\,\underline\xi_k(\chi)
\end{equation*}
where $\alpha_k\ge0$ is the task-dependent weight of capability $k$.
Let $C_{\rm acq}(\chi|q_0)$ denote the acquisition cost of transitioning from the robot's current whole-body configuration $q_0$ to candidate $\chi$, and let $\alpha_{\rm acq}>0$ be its weight.
The contact and whole-body configuration selection problem is
\begin{equation}
\begin{aligned}
\chi^\star
\in
\operatorname*{arg\,max}_{\chi}
\quad
& J(\chi)=U(\underline\xi(\chi))-\alpha_{\rm acq} C_{\rm acq}(\chi|q_0)\\
\mathrm{s.t.}\quad
& \text{\textbf{(C1)}, \textbf{(C2)}, and \textbf{(T1)}} .
\end{aligned}
\label{eq:contact_selection}
\end{equation}
Conditions 
\textbf{(C1)} and \textbf{(C2)}  
ensure that the environmental contact is realized and the contact mode is maintained.
Condition 
\textbf{(T1)} 
ensures that the required task values are 
supported in every stance phase. 
Among feasible candidates, the objective $J(\chi)$ favors greater residual
capabilities and lower acquisition cost.

\noindent\textbf{Challenges.}
Conditions \textbf{(C1)} and \textbf{(C2)} can be checked directly through forward kinematics, geometry, and contact-point velocity. In contrast, 
evaluating $J(\chi)$ for every candidate is computationally expensive, as we need to solve many whole-body optimization problems. This number also grows with the candidate set, which may need to be large to explore the high-dimensional configuration space across the environmental surfaces.
Moreover, nearby candidates may have different active torque, friction, and contact constraints, so their capabilities can vary nonsmoothly.
Therefore, evaluating every candidate is costly, while a single global approximation of the capability map can be unreliable.

\section{Methodology}
\label{sec:approach}
Evaluating the capability vector $\xi(\chi)$ exactly for every candidate requires solving many whole-body optimization problems across the task-applicable capabilities and stance phases. To address this, we propose \textit{Capability-Tradeoff Contact Selection} (CTCS), which combines exact feasibility screening, local
sensitivity-based capability prediction, and selective exact evaluation.

As visualized in Fig.~\ref{fig:overview}, CTCS includes five stages. 
We first generate a finite set of candidate contact realizations and screen them for contact realization and task feasibility. 
Then, we categorize the surviving candidates into groups according to their whole-body realizations and environmental-contact Jacobians. For each group, one representative candidate is evaluated and used as an anchor to predict the capabilities of nearby candidates through local sensitivity analysis.
These predictions are validated through selected exact evaluations. Finally, candidates are ranked using their estimated capabilities, and the top-ranked candidates are evaluated exactly before the final selection is made according to~\eqref{eq:contact_selection}.

\subsection{Candidate Generation and Task Feasibility Screening}
\label{subsec:generation}

The search space of robot configurations is continuous and high-dimensional, so
we construct a finite set of candidate configurations through sampling. 
For each surface $\Pi\in\mathcal E$, we sample changes from the current configuration in the manipulator joint angles, the base position in the two directions orthogonal to $n_\Pi$, the base roll and pitch, and the four foothold locations, uniformly within bounds. We collect these sampled quantities in $\theta$. 

The base position follows from the contact condition \textbf{(C1)} and the leg joint angles follow from the foothold inverse kinematics. Together with the sampled manipulator joint angles, these realizations give the whole-body configuration $q(\theta)$ (explicit form in the Supplementary Material~\cite{supp}). The sampling bounds keep the legs away from kinematic singularities, including full extension, and the manipulator away from singular configurations, so that the realization equations remain locally nonsingular and $q$ depends differentiably on $\theta$. We then filter out the configurations that do not satisfy \textbf{(C1)} and \textbf{(C2)},  and continue sampling until $N$ candidates are retained. The resulting candidate set is $\mathcal X_N=\{\chi_j=(q_j,\Pi_j)\}_{j=1}^{N}$, 
where $q_j$ is the whole-body configuration and $\Pi_j\in\mathcal E$ is the environmental surface with which that configuration is in contact.

Next, we screen the candidates for the prescribed task requirements.
For every candidate $\chi_j$ and stance phase $s\in\mathcal S$, we evaluate the task-feasibility condition \textbf{(T1)}. 
We use $\mathcal D_1\subseteq\mathcal X_N$ to denote the set of candidates satisfying \textbf{(T1)} in every stance phase. Only candidates in $\mathcal D_1$ are considered in the subsequent processes.

\begin{figure}[t]
\centering
\includegraphics[width=0.42\textwidth]{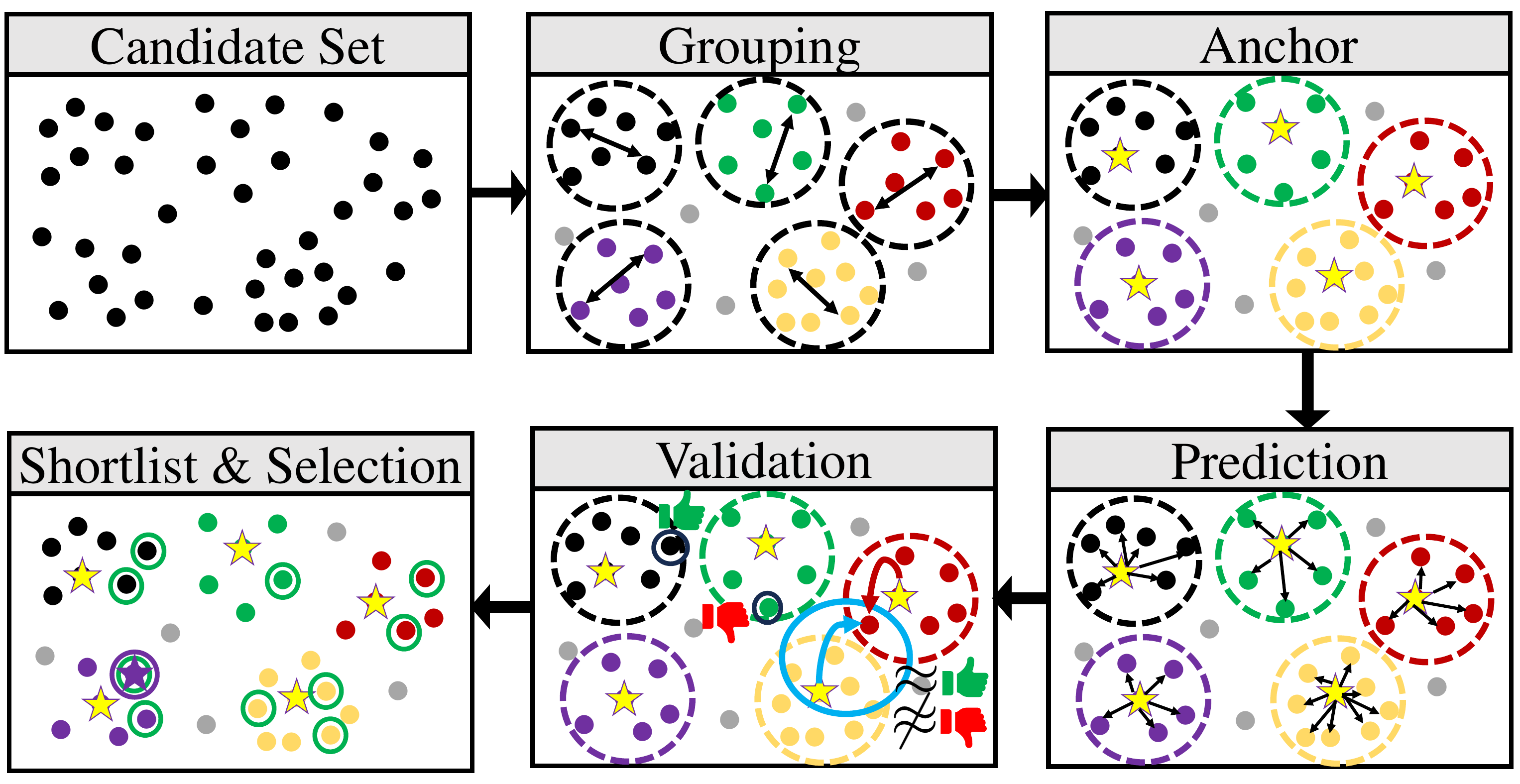}
\caption{\small {Overview of CTCS: candidate grouping, anchor selection, group members' capability prediction, validation, and final selection. Yellow stars are anchors and the purple star is the selected candidate.\vspace{-1.5em}}
}
\label{fig:overview}
\end{figure}

\subsection{Grouping and Anchor Selection}
\label{subsec:anchors}
Evaluating the surviving candidates in $\mathcal D_{1}$ exactly still requires solving all task-applicable directional capability optimization problems in every stance phase, which can be computationally costly. To reduce this cost, we seek to group candidates with locally similar capability optimization problems, so that an exact evaluation of one representative candidate can be used to predict the capabilities of other members in the group. To this end, grouping directly using contact surface is not sufficient, since even on the same surface, different whole-body realizations can induce different actuation, friction, and active constraints.

\noindent\textbf{Grouping.}
To address the above challenge, we propose to construct the groups using both realization coordinate $\theta$ and environmental-contact Jacobian $J_c(q)$. Here, $\theta$ captures configuration-dependent changes in the robot dynamics and constraints. Matrix $J_c(q)$ additionally characterizes how the contact forces impact the whole-body dynamics and how maintaining the contact constrains robot motion. We define: 
\begin{equation}
\begin{aligned}
d^2(\chi_i,\chi_j)=\|\theta_i-\theta_j\|_{W_\theta}^2 +w_J
\| J_c(q_i)- J_c(q_j)\|_F^2,
\end{aligned}
\label{eq:group_metric}
\end{equation}
where $W_\theta\succ0$ and $w_J>0$ are weights that balance the two terms, and $\|\cdot\|_F$ denotes the Frobenius norm.
We form groups on each surface by complete-linkage clustering~\cite{hansen1978complete} 
with a maximum group diameter $\Delta_{\rm grp}$, which bounds within-group differences in realization and contact mappings.

\noindent\textbf{Anchor Selection.}
A first-order prediction requires an exactly evaluated expansion point with reliable local sensitivity.
For each member $i$ within a group $\mathcal G$, we measure its maximum distance to other members as $d_{\max}(\chi_i)
=
\max_{\chi_j\in\mathcal G} d(\chi_i,\chi_j).$
We order the candidates in ascending order of $d_{\max}(\chi_i)$ and test them in this order. Thus, candidates that are more centrally located within the group are considered first. For each tested candidate, we exactly solve all task-applicable directional capability optimization problems in every stance phase. The candidate is accepted as an anchor only if the resulting solutions satisfy the following two conditions:
\textbf{(R1)} Every directional LP has a nondegenerate optimal basis with margin $\tau_{\rm nd}>0$, that is, each basic variable is at least $\tau_{\rm nd}$ away from its bounds, and each nonbasic reduced cost is at least $\tau_{\rm nd}$ in magnitude. This gives a locally stable optimal basis and a well-defined value sensitivity. 
\textbf{(R2)} Every directional reserve $r_{k,j}^s$ of the task-applicable capability measures at that member is at least a capability-specific floor $r_{{\rm min},k}>0$.
This avoids capability estimates being dominated by near-zero reserves.

The first member of the order that satisfies \textbf{(R1)}--\textbf{(R2)} becomes the anchor of its group. If no member satisfies them, we do not construct a local model for that group and instead evaluate all of its members exactly.

\subsection{Anchor-based Prediction, Validation, and Selection}
\label{subsec:sensitivity}

The grouping criterion and anchor conditions identify regions where local capability prediction is likely to be reliable, but they do not guarantee that the same local model remains accurate throughout a group. We construct a first-order prediction from each anchor, then perform two validity checks on the predictions: (i) a probe test and (ii) a cross-anchor check. 

\noindent\textbf{Anchor Sensitivity-based Prediction.}
Let $\hat\chi=(\hat q,\Pi)$ be an anchor of a group with coordinate $\hat\theta$.
The realization equations define $q$ locally as a differentiable function
of $\theta$. Denote its sensitivity at the anchor by $Y = D_{\theta}q\big|_{\hat{\theta}}$, obtained from the realization equations through the implicit
function theorem~\cite{ha2017joint} (derivation in the Supplementary Material~\cite{supp}).

Let $p$ index the task-applicable directional optimization problems of a stance phase. For each $p$ and each phase $s$, let $\mathcal V_{p,s}$ and $\mathcal L_{p,s}$ be the optimal value and the Lagrangian of that problem at the anchor.
Condition \textbf{(R1)} ensures local regularity, so one has $\nabla_\theta\mathcal V_{p,s}(\hat\chi)=Y^\top\nabla_q\mathcal L_{p,s}(\hat\chi),$
where $\nabla_q\mathcal L_{p,s}(\hat\chi)$ differentiates the function at the fixed optimal primal-dual pair.
For a member $\chi_i$ of that group with displacement $\delta\theta_i=\theta_i-\hat\theta$, we first form the linear prediction
\vspace{-0.4em} 
\begin{equation}
\widetilde{\mathcal V}_{p,s}(\chi_i)
=
\mathcal V_{p,s}(\hat\chi)
+
\nabla_\theta\mathcal V_{p,s}^{\top}\delta\theta_i .
\label{eq:local_prediction}
\end{equation}
Because the directional optimization values represent nonnegative reserve magnitudes, we use 
$\widehat{\mathcal V}_{p,s}(\chi_i) =
\max\big\{0,\widetilde{\mathcal V}_{p,s}(\chi_i)\big\}$.
We predict all task-applicable capability reserves unless their exact values are already available.
We then build the capabilities from these values, using the exact values when available and, otherwise, the predicted values by taking the minimum of each positive-negative pair to obtain $r_{k,j}^s$, and then computing 
$\rho_{\rm wrench}^s$, $\rho_{\rm reach}^s$, and $\rho_{\rm mob}^s$. Finally, we take the minimum over stance phases to obtain each component of the predicted capability vector $\hat{\underline\xi}(\chi_i)$.

\noindent\textbf{Prediction Validation.}
We validate each local model by exactly evaluating the group member farthest from its anchor using~\eqref{eq:group_metric}. If the prediction error of any task-applicable capability exceeds \(\epsilon_{\mathrm{pr},k}\), we regroup using maximum diameter \(\Delta_{\mathrm{grp}}/2\) and repeat the probe once. Any resulting group that fails again is evaluated exactly.

We further perform a \emph{cross-anchor check}: A member that lies within $\Delta_{\rm grp}$ under~\eqref{eq:group_metric} of another group's valid anchor on the same surface receives a second prediction from that anchor. If the two predictions differ on any capability $k$ by more than the capability-specific cross-anchor tolerance $\epsilon_{\times,k}>0$, the member is evaluated exactly.

\noindent\textbf{Ranking and Selection.}
To rank the candidates, we estimate their capability utility as
$$\textstyle\widehat{J}(\chi)=\sum_{k\in\mathcal K_{\rm task}}\alpha_k\hat{\underline\xi}_k(\chi)-\alpha_{\rm acq} C_{\rm acq}(\chi|q_0),$$
where $\hat{\underline\xi}_k(\chi)$ is the prediction of capability $k$ (or exact value when available) and $C_{\rm acq}(\chi|q_0)$ is the acquisition cost.
We rank the candidates by decreasing $\widehat{J}(\chi)$ and select up to $N_{\rm short}$ top-ranked candidates for exact evaluation and denote the set by $\mathcal D_{\rm short}$. For the shortlisted candidates, we solve the remaining task-applicable directional optimization problems, reusing any available exact results, so that all of their capability measures are exact.
We then select
\vspace{-0.5em}
\begin{equation}
\chi^\star
\in
\operatorname*{arg\,max}_{\chi\in\mathcal D_{\rm short}}
J(\chi).
\label{eq:exact_select}
\end{equation}
Every candidate in $\mathcal D_{\rm short}$ satisfies \textbf{(C1)}, \textbf{(C2)}, and \textbf{(T1)} and has exactly computed capabilities for~\eqref{eq:contact_selection}.

\section{Experiments and Results}
\label{sec:results}

We evaluate CTCS on a quadruped mobile manipulator in environments where multiple support surfaces are available. 
We assess the contact selection quality, computational cost, and task performance, and compare them with baselines. The validation is performed on both Gazebo simulations and hardware experiments.

\noindent\textbf{Hardware and Environment Setup.}
We use a Unitree Go2 quadruped equipped with an AgileX NERO 7-DoF manipulator and a rigid planar support frame mounted on the side of its body for establishing environmental contact (Fig.~\ref{fig:concept_robot_pic}).
We consider $9$ different support surfaces simultaneously ($|\mathcal E| = 9$), each $5\,\mathrm{m}$ long:
one vertical wall, 
four tables and four horizontal rails at different heights.
For each task condition, the robot is not assigned a 
support; 
it selects among all nine surfaces and multiple whole-body configurations that realize contact with each surface.
Fig.~\ref{fig:gazebo_env} shows representative contact realizations on four of these surfaces.

\begin{figure}[t]
\centering
\includegraphics[width=.8\columnwidth]{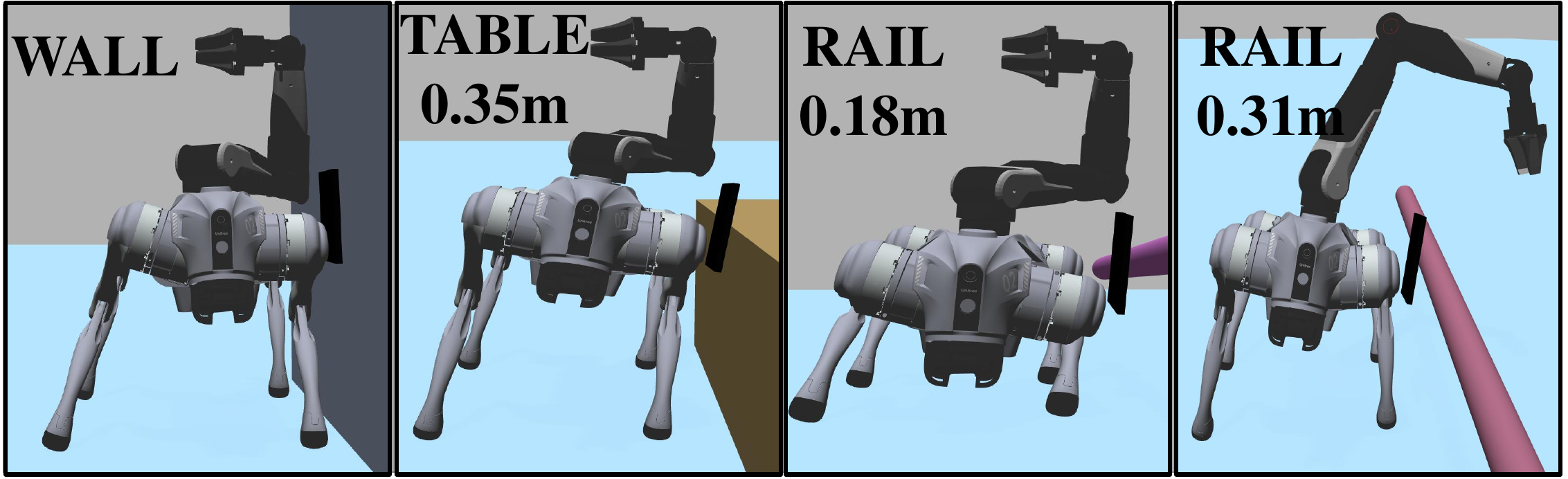}
\caption{\small Examples of different environmental surfaces and the corresponding contact realization of the mobile manipulator.\vspace{-2.0em}}
\label{fig:gazebo_env}
\end{figure}

\noindent\textbf{Tasks and Conditions.} 
We evaluate both sticking and walking modes (with two diagonally supported trot phases). 
For each task condition, we choose the desired end-effector position, and for walking mode, we set the required base velocity at $0.1\,\mathrm{m/s}$ along the support surface.
For each mode, we vary the force component of the required end-effector wrench over $14$ force directions and $14$ force magnitudes ranging from $5$ to $70\mathrm{N}$. 
This gives $196$ conditions per mode and $2\times14\times14 = 392$ task conditions in total.
Task requirement quantities are given in the Supplementary Material~\cite{supp}.

\noindent\textbf{Candidate Generation.} For each surface, we sample configurations and keep those satisfying \textbf{(C1)} and \textbf{(C2)} until $5000$ candidates remain. Each candidate set therefore contains $45000$ candidates across the $9$ surfaces.
Averaged over the $392$ task conditions, $27851$ of these candidates satisfy \textbf{(T1)}.

\noindent\textbf{Control Implementation.} 
Selected configurations are executed using a standard receding-horizon
whole-body controller that tracks the selected configuration and
prescribed motion while enforcing the whole-body dynamics and contact
constraints. At each control step, only the first optimized command is
applied before replanning. Additional controller details are provided
in the Supplementary Material~\cite{supp}.

\noindent\textbf{Baselines and Ablations.}
We compare CTCS with three baselines: (i) \textit{Exhaustive}, which exactly evaluates all candidates in $\mathcal D_1$ (the set satisfying constraints);
(ii) \textit{Ground-only}, which evaluates the whole-body configurations with no environmental surfaces considered; and
(iii) \textit{Fixed-Contact}, which evaluates one prescribed candidate, with a fixed whole-body configuration, support surface, and contact location.

We evaluate ablations: (i) \textit{CTCS-NV}, which removes prediction validation while retaining final evaluation; (ii) \textit{CTCS-SA}, which removes grouping but instead uses a single anchor per surface;
and (iii) \textit{CTCS-FFS} for walking tasks, which selects candidates using only the four-foot-stance model and then evaluates its feasibility in diagonal trot phases.

\subsection{Environmental Contact Selection}
\label{subsec:env_contact_selec}

\begin{table}[t]
\centering
\scriptsize
\setlength{\tabcolsep}{2.2pt}
\renewcommand{\arraystretch}{1.05}
\caption{\small Gazebo simulations: Evaluation over $196$ task conditions per mode.
Columns report certification rate, mean objective over all $196$ conditions, with zero assigned when no candidate is certified, selection agreement with exhaustive search over the $180$ certifiable conditions, runtime, and runtime speedup relative to exhaustive search.
$\Uparrow$ indicates higher is better and $\Downarrow$ indicates lower is better.}
\label{tab:main}

\resizebox{\columnwidth}{!}{%
\begin{tabular}{@{}lllccccc@{}}
\toprule
Mode & Type & Method & Certification & Mean objective
& Selection & Runtime & Speedup \\
& & & rate (\%) $\Uparrow$ & $\Uparrow$
& agreement (\%) $\Uparrow$ & (s) $\Downarrow$ & ($\times$) $\Uparrow$ \\
\midrule

\multirow{7}{*}{Sticking}
& \multirow{3}{*}{Baseline}
& Exhaustive     & 91.8 & 3.5428 & 100.0 & 124.0 & 1.00 \\
& & Ground-only   & 68.4 & 2.5324 & 64.4  & 83.2  & 1.49 \\
& & Fixed-Contact & 48.5 & 2.1386 & 0.0   & 8.3   & 14.89 \\
\cmidrule(lr){2-8}
& Proposed
& \textbf{CTCS} & 91.8 & 3.5419 & 95.6 & 39.1 & 3.17 \\
\cmidrule(lr){2-8}
& \multirow{3}{*}{Ablation}
& CTCS-NV & 91.8 & 3.4318 & 92.2 & 29.5 & 4.21 \\
& & CTCS-SA & 91.8 & 3.1687 & 85.6 & 25.2 & 4.92 \\

\midrule
\midrule

\multirow{8}{*}{Walking}
& \multirow{3}{*}{Baseline}
& Exhaustive     & 91.8 & 5.6612 & 100.0 & 157.0 & 1.00 \\
& & Ground-only   & 65.3 & 3.9184 & 2.8   & 78.1  & 2.01 \\
& & Fixed-Contact & 41.8 & 2.0812 & 0.0   & 11.8  & 13.34 \\
\cmidrule(lr){2-8}
& Proposed
& \textbf{CTCS} & 91.8 & 5.6612 & 100.0 & 62.1 & 2.53 \\
\cmidrule(lr){2-8}
& \multirow{4}{*}{Ablation}
& CTCS-NV  & 91.8 & 5.4873 & 95.0 & 54.0 & 2.91 \\
& & CTCS-SA  & 91.8 & 5.3982 & 92.8 & 46.3 & 3.39 \\
& & CTCS-FFS & 91.8 & 3.0387 & 2.8  & 63.6 & 2.47 \\

\bottomrule
\end{tabular}%
}%
\vspace{-2.3em}
\end{table}

The quantitative evaluations in Table~\ref{tab:main} are from Gazebo simulations over the $392$ task conditions.
It shows how CTCS improves both certification rate (fraction of task conditions for which a feasible candidate is returned) and mean objective relative to ground-only support.
In sticking mode, CTCS certifies $180$ of the $196$ conditions ($91.8\%$), compared with $134$ ($68.4\%$) for ground-only search.
In walking mode, the corresponding counts are $180$ ($91.8\%$) and $128$ ($65.3\%$).
Since ground-only search evaluates the same sampled whole-body configurations, these results show that the additional environmental support helps the robot satisfy the task requirements that ground support alone cannot accommodate with this candidate set.
The mean objective also increases from $2.5324$ to $3.5419$ in sticking mode and from $3.9184$ to $5.6612$ in walking mode, respectively. Since uncertified conditions are assigned zero in the reported mean, these improvements reflect both certification coverage and the quality of the selected candidates.

However, using an environmental contact does not by itself ensure better performance. 
The fixed-contact baseline certifies only $95$ ($48.5\%$) sticking conditions and $82$ ($41.8\%$) walking conditions, with mean objectives of $2.1386$ and $2.0812$, respectively. It therefore performs worse than ground-only search on both metrics. Fixing the candidate also fixes the whole-body configurations and contact geometry, which may not suit a given task.
For the prescribed candidate tested here, these results show the importance of selecting the contact and configuration according to the task requirements.

\subsection{Selection Quality and Computational Cost}
\label{subsec:quality}

\noindent\textbf{Comparison with Exhaustive Sample Evaluation.}
As shown in Table~\ref{tab:main}, CTCS reduces runtime from $124.0\,\mathrm{s}$ to $39.1\,\mathrm{s}$ in sticking mode and from $157.0\,\mathrm{s}$ to $62.1\,\mathrm{s}$
in walking mode, corresponding to $3.17\times$ and $2.53\times$ speedups.
Despite evaluating only a subset of candidate capabilities exactly,
CTCS closely matches exhaustive evaluation: its selection agreement is
95.6\% in sticking mode and 100\% in walking mode, while the mean
objective changes from 3.5428 to 3.5419 in sticking mode and is
unchanged at 5.6612 in walking mode. Thus, the candidates selected
differ occasionally, but these differences have negligible effect on
the aggregate objective in these experiments.
\label{subsec:structures}
\begin{figure}[t]
\centering
\includegraphics[width=.9\columnwidth]{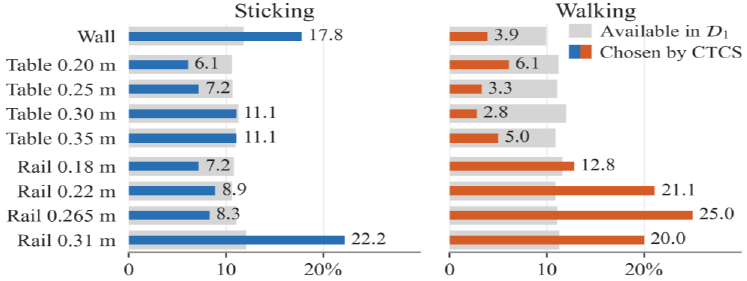}
\caption{\small Surface availability and CTCS selection rates over $180$ certified conditions per mode. All values are percentages; heights are in meters.\vspace{-2.0em}}
\label{fig:structures}
\end{figure}

\noindent\textbf{Ablations.}
Removing prediction validation (CTCS-NV) or replacing the local groups
with a single anchor per surface (CTCS-SA) further reduces runtime but
also lowers the mean objective.
The former indicates that the probe and cross-anchor checks help detect
inaccurate predictions before the final selection. 
The latter supports the effectiveness of grouping candidates within each surface and using multiple local anchors, since candidates on the same surface can have
different whole-body configurations and limiting constraints.
Finally, evaluating walking candidates using a four-foot stance
(CTCS-FFS) reduces the mean objective to 3.0387 despite using the same
task-feasible candidate set. This shows that selection should account for
the residual capabilities under the actual trot support phases.

\subsection{Task-Dependent Environmental Structure Selection}
Fig.~\ref{fig:structures} compares each surface's share of feasible candidates in $\mathcal D_1$ with its CTCS selection rate. Although candidate configurations are sampled uniformly for each surface
before screening, their availability afterward depends on task feasibility. In our experiments, this availability remains similar across surfaces, ranging from $10.6\%$ to $12.1\%$ in sticking mode and from $9.9\%$ to $12.0\%$ in walking mode, while selection rates differ substantially. Thus, similar availability does not imply equally favorable candidates for the task.

In walking mode, CTCS selects rails in $142$ of the $180$ certified conditions ($78.9\%$), most often the $0.265\,\mathrm{m}$ rail ($25.0\%$). In contrast, the wall is selected in only $3.9\%$ of conditions.
One possible reason is that the rails' lower modeled kinetic friction coefficient can reduce sliding friction, depending on the contact normal forces. This can make rails more favorable for walking.

In sticking mode, the selections are more broadly distributed.
The $0.31\,\mathrm{m}$ rail is selected most often ($22.2\%$), followed by the wall ($17.8\%$). The wall's greater use than in walking is consistent with its higher modeled static friction coefficient that resists sliding. However, surface height also affects the whole-body configuration, joint effort, and margins to physical limits. This may contribute to the preference for the $0.31\,\mathrm{m}$ rail over the other rails, despite their shared friction coefficients.
These results show that the preferred environmental support depends jointly on contact properties, support geometry, and the task requirements.

\begin{figure}[t]
\centering
\includegraphics[width=.9\columnwidth]{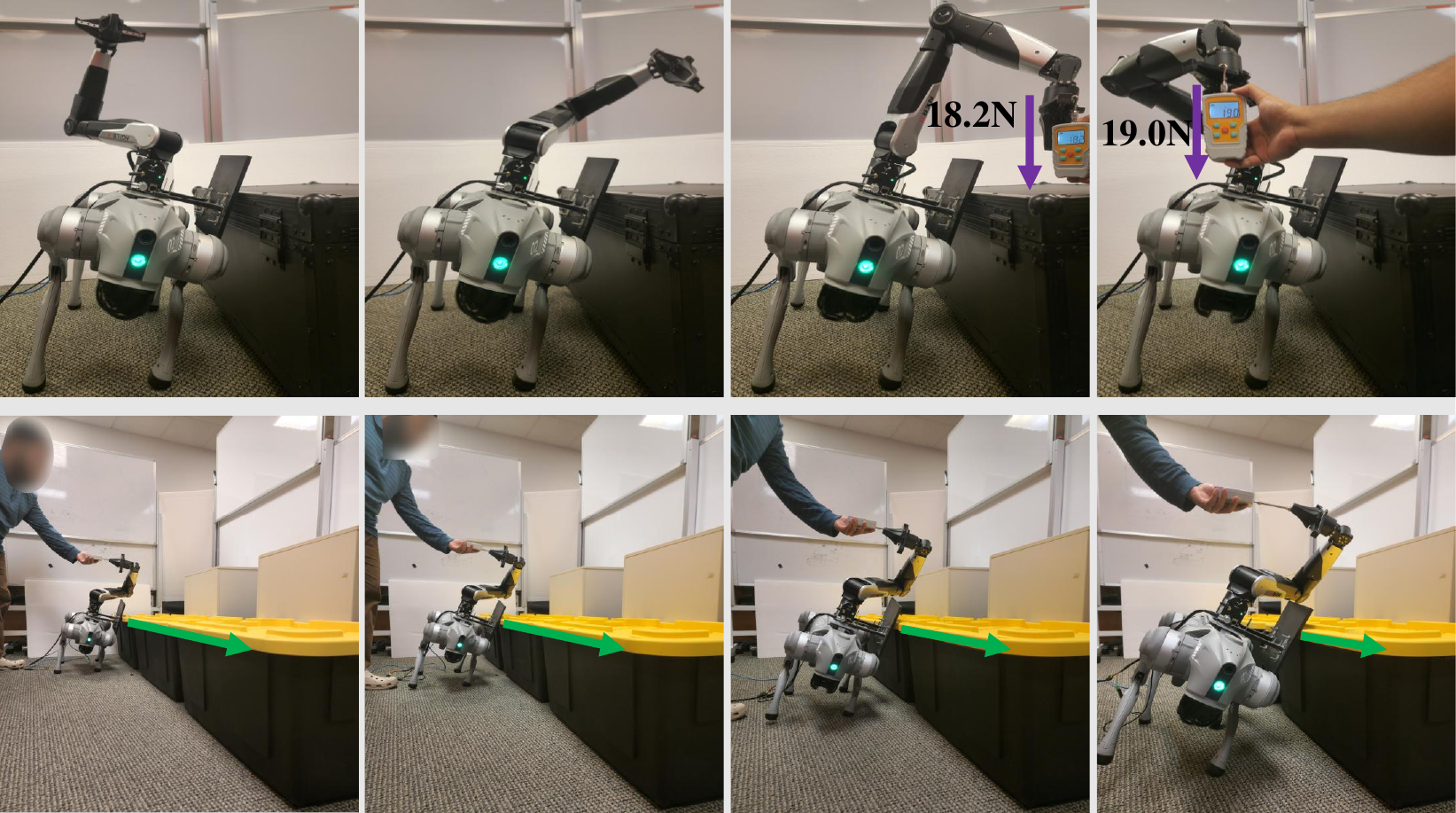}
\caption{\small Hardware experiment. {Upper: A sticking-mode manipulation task with an applied downward force while maintaining selected support contact.}
{Lower: A walking task. The robot sustains the applied force while maintaining the sliding support contact, closely tracking the required base velocity and end-effector position.}
\vspace{-2.0em}
}
\label{fig:manip_seq}
\end{figure}

\subsection{Hardware Experiments}
\label{subsec:hardware_exp}
We further validate the contact realizations selected by CTCS through hardware experiments with different environmental surfaces. The receding-horizon controller runs at $10\,\mathrm{Hz}$. Table~\ref{tab:tracking} reports the tracking errors averaged over five hardware trials for both sticking and walking tasks. 

Fig.~\ref{fig:manip_seq}-Upper shows a representative sticking-mode manipulation task in which the robot selects a contact and whole-body configuration and the end-effector moves from an initial pose toward a target pose while maintaining environmental support contact. During the task, we apply a downward load, and the end-effector sustains it. Over five trials, the force and end-effector position RMSEs remain low (Table~\ref{tab:tracking}-Left). 
Without environmental support, due to the weight of the arm, the robot will roll over. 

\begin{table}[b]
\centering
\caption{\small Hardware trials tracking errors.
Force RMSE: force-magnitude root mean squared error;
Vel. ME: mean base velocity error;
EP RMSE: end-effector position root mean squared error.}
\label{tab:tracking}

\begin{minipage}[t]{0.40\columnwidth}
\centering
\scriptsize
\textbf{Sticking-mode Task}\par
\vspace{1.5pt}
\setlength{\tabcolsep}{1pt}
\begin{tabular*}{\linewidth}{@{\extracolsep{\fill}}cc@{}}
\toprule
Force RMSE & EP RMSE \\
\midrule
$0.229$ $\mathrm{N}$ & $6.48$ $\mathrm{mm}$ \\
\bottomrule
\end{tabular*}
\end{minipage}%
\hspace{0.025\columnwidth}%
\begin{minipage}[t]{0.54\columnwidth}
\centering
\scriptsize
\textbf{Walking-mode Task}\par
\vspace{1.5pt}
\setlength{\tabcolsep}{1pt}
\begin{tabular*}{\linewidth}{@{\extracolsep{\fill}}ccc@{}}
\toprule
Force RMSE & Vel.\ ME & EP RMSE \\
\midrule
$0.278$ $\mathrm{N}$ & $-0.00547$ $\mathrm{m/s}$ & $14.97$ $\mathrm{mm}$ \\
\bottomrule
\end{tabular*}
\end{minipage}
\end{table}

Fig.~\ref{fig:manip_seq}-Lower shows a complete walking experiment using a 
contact realization selected by CTCS. 
The robot walks along the surface while maintaining environmental contact and sustaining the applied load, with the prescribed base velocity and end-effector position as tracking references. Across five trials, the force, base-velocity, and
end-effector position errors remain low (Table~\ref{tab:tracking}-Right). The larger end-effector position error than in sticking mode is due to movement of the body.
This experiment also illustrates capability trade-offs: the additional contact increases the wrench that the robot can sustain, while the robot must overcome the corresponding frictional resistance during locomotion.

Fig.~\ref{fig:error_fig} further shows a representative walking trial.
The measured wrist-reaction force follows the applied load, while the
base velocity fluctuates around the desired $0.1\,\mathrm{m/s}$ and the
end-effector position error generally remains below $30\,\mathrm{mm}$.
The remaining fluctuations are mainly due to the alternating ground
support and body motion during trot changes.

\begin{figure}[t]
\centering
\includegraphics[width=0.9\columnwidth]{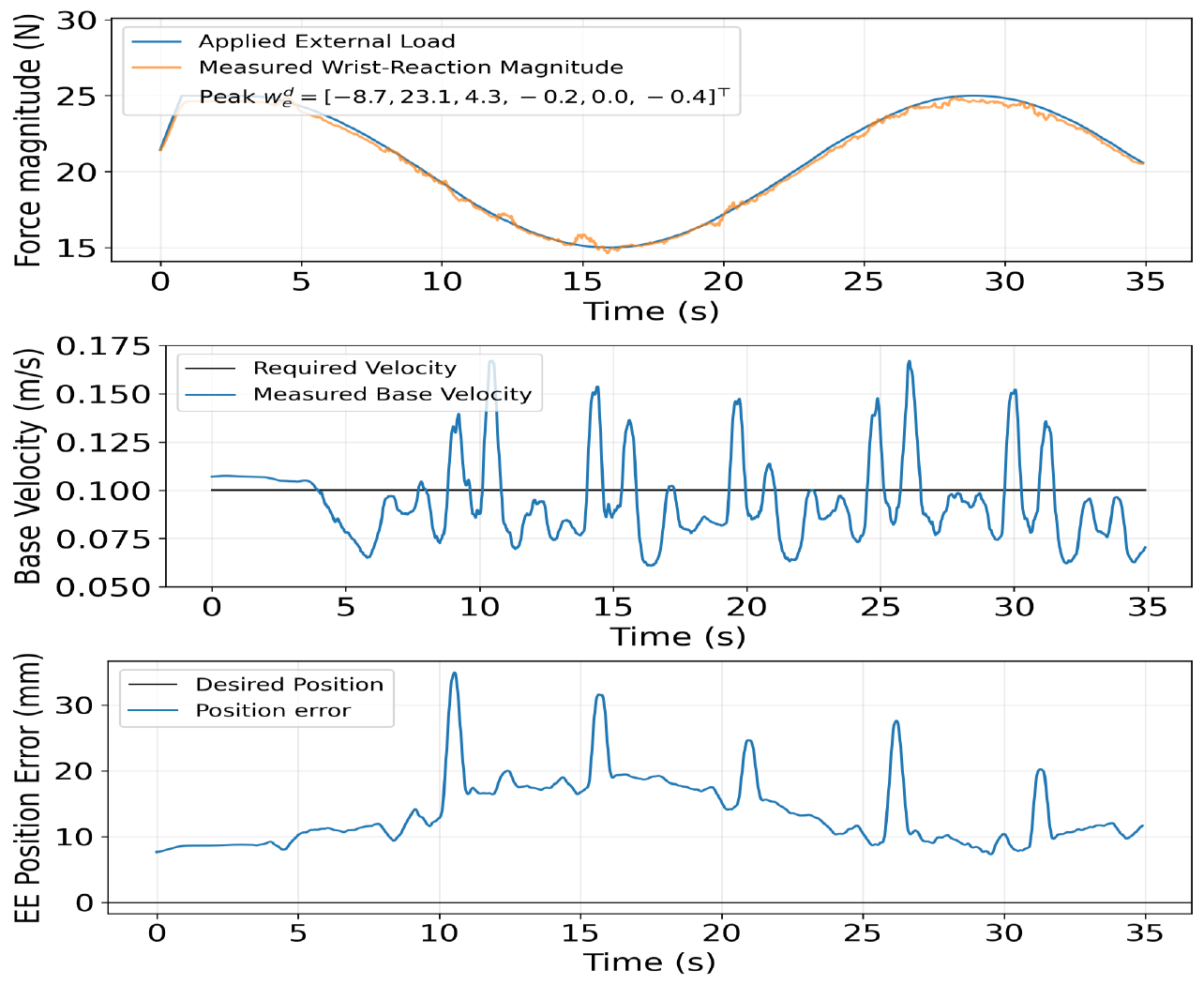}
\caption{\small{Task performance in a representative hardware walking experiment. The peak of applied force magnitude is $25~\mathrm{N}$, required base velocity is $0.1\,\mathrm{m/s}$ with the desired end-effector pose $[0.10,\,0.15,\,0.55]^\top\,\mathrm{m}$ during walking. We record the data at $10\,\mathrm{Hz}$.\vspace{-2.0em}}}
\label{fig:error_fig}
\end{figure}

\section{Conclusion}
\label{sec:conclusion}
In this paper, we presented CTCS to jointly select environmental support contacts and whole-body configurations for prescribed loco-manipulation tasks. Our results showed that the benefit of environmental contact depends on both the selected surface and configuration, with task-dependent selection outperforming the tested fixed-contact baseline.
CTCS closely matched the mean objective of exhaustive evaluation with less computation, while the ablations showed that local grouping and prediction checks improved selection quality.
These results support treating environmental contact as a task-dependent decision.
One limitation of our approach is that it selects from sampled candidates, so global optimality over the continuous space of contact locations and configurations is not guaranteed.
Future work will include candidate refinement, contact transitions between surfaces, and online reselection as task requirements change.

\bibliography{bibliography}
\bibliographystyle{ieeetr}

\end{document}